\documentclass[letterpaper]{article} % DO NOT CHANGE THIS
\usepackage{aaai25}  % DO NOT CHANGE THIS
\usepackage{times}  % DO NOT CHANGE THIS
\usepackage{helvet}  % DO NOT CHANGE THIS
\usepackage{courier}  % DO NOT CHANGE THIS
\usepackage[hyphens]{url}  % DO NOT CHANGE THIS
\usepackage{graphicx} % DO NOT CHANGE THIS
\usepackage{natbib}  % DO NOT CHANGE THIS AND DO NOT ADD ANY OPTIONS TO IT
\usepackage{caption} % DO NOT CHANGE THIS AND DO NOT ADD ANY OPTIONS TO IT
\usepackage{algorithm}
\usepackage{algorithmic}

\usepackage{newfloat}
\usepackage{listings}
\DeclareCaptionStyle{ruled}{labelfont=normalfont,labelsep=colon,strut=off} % DO NOT CHANGE THIS
\floatstyle{ruled}
\newfloat{listing}{tb}{lst}{}
\floatname{listing}{Listing}
\title{Physics-Guided Fair Graph Sampling for Water Temperature \\ Prediction in River Networks}
\author{
    Erhu He\textsuperscript{\rm 1},
    Declan Kutscher\textsuperscript{\rm 1},
    Yiqun Xie\textsuperscript{\rm 2},
    Jacob Zwart\textsuperscript{\rm 3}, 
    Zhe Jiang\textsuperscript{\rm 4},
    Huaxiu Yao\textsuperscript{\rm 5},
    Xiaowei Jia\textsuperscript{\rm 1}
}
\affiliations{
    \textsuperscript{\rm 1}University of Pittsburgh\\
    \textsuperscript{\rm 2}University of Maryland\\
    \textsuperscript{\rm 3}U.S. Geological Survey\\
    \textsuperscript{\rm 4}University of Florida\\
    \textsuperscript{\rm 5}University of North Carolina at Chapel Hill\\
    
\textsuperscript{\rm 1}\{erh108, dtk28, xiaowei\}@pitt.edu,
\textsuperscript{\rm 2}xie@umd.edu,
\textsuperscript{\rm 3}jzwart@usgs.gov,
\textsuperscript{\rm 4}zhe.jiang@ufl.edu,
\textsuperscript{\rm 5}huaxiu@cs.unc.edu
}

\usepackage{bibentry}
\usepackage{algorithm}
\usepackage{algorithmic}
\usepackage{booktabs}
\usepackage{amsmath}
\usepackage{graphicx}
\usepackage{subfigure}
\usepackage{amsfonts,bm}
\usepackage{multirow}
\usepackage{color}
\usepackage{braket}
\usepackage{mathtools}
\usepackage{enumerate}
\usepackage{enumitem}
\usepackage{empheq}
\usepackage{calc}
\usepackage{dsfont}
\usepackage{url}
\usepackage[flushmargin]{footmisc}%hang,

\usepackage[flushleft]{threeparttable}

\usepackage{placeins}

\usepackage{algorithm}
\usepackage{algorithmic}

\floatstyle{ruled}
\newfloat{listing}{tb}{lst}{}
\floatname{listing}{Listing}

\newcommand{\mf}{\mathcal{F}}

\newcommand{\p}{\mathcal{P}}

\newcommand{\vtheta}{\mathbf{\Theta}}

\begin{document}

\maketitle

\begin{abstract}
This work introduces a novel graph neural networks (GNNs)-based method to predict stream water temperature and reduce model bias across locations of different income and education levels. Traditional physics-based models often have limited accuracy because they are necessarily approximations of reality. Recently, there has been an increasing interest of using GNNs in modeling complex water dynamics in stream networks. Despite their promise in improving the accuracy, GNNs can bring additional model bias through the aggregation process, where node features are updated by aggregating neighboring nodes. The bias can be especially pronounced when nodes with similar sensitive attributes are frequently connected. We introduce a new method that leverages physical knowledge to represent the node influence in GNNs, and then utilizes physics-based influence to refine the selection and weights over the neighbors. The objective is to facilitate equitable treatment over different sensitive groups in the graph aggregation, which helps reduce spatial bias over locations, especially for those in underprivileged groups. The results on the Delaware River Basin demonstrate the effectiveness of the proposed method in preserving equitable performance across locations in different sensitive groups. 
\end{abstract}

\section{Introduction}
Water quality continues to degrade due to the increasing demand on water-related services (e.g.,  drinking water supply and energy consumption) and the changing climate~\cite{hoekstra_water_2012, carr_recent_2013}. Timely monitoring of water temperature, which serves as a master factor of water quality,  can help  inform water management decisions (e.g., water filtering, water release from reservoirs, enhanced management of nearby farmlands), maintain the desired habitat for aquatic life, and 
better understand ecological processes leading to the changes in water quality.

To address this problem, scientists from multiple disciplines have built physics-based models
to simulate the internal water cycles and their interactions with weather, soil, and other water bodies based on general physical relationships such as energy and mass conservation~\cite{hipsey2014glm,markstrom2012p2s,dugdale2017river}. However, these models still rely on qualitative parameterizations (i.e., approximations) based on soil and surficial geologic classification along with topography, land cover, and climate input, leading to degraded predictive accuracy. 
Recently, graph neural networks (GNNs) have been widely adopted as a data-driven solution to model stream networks~\cite{jia2021physics_sdm,moshe2020hydronets,sun2021explore,topp2023stream,chen2022physics,jia2021physics_simlr,chen2021heterogeneous,jia2023physics} because they can learn to capture complex interactions amongst different stream segments (e.g., through mass advection and diffusion). 

\begin{figure} [!t]
\centering
\subfigure[]{ \label{fig:mot1}{}
\includegraphics[width=0.475\columnwidth]{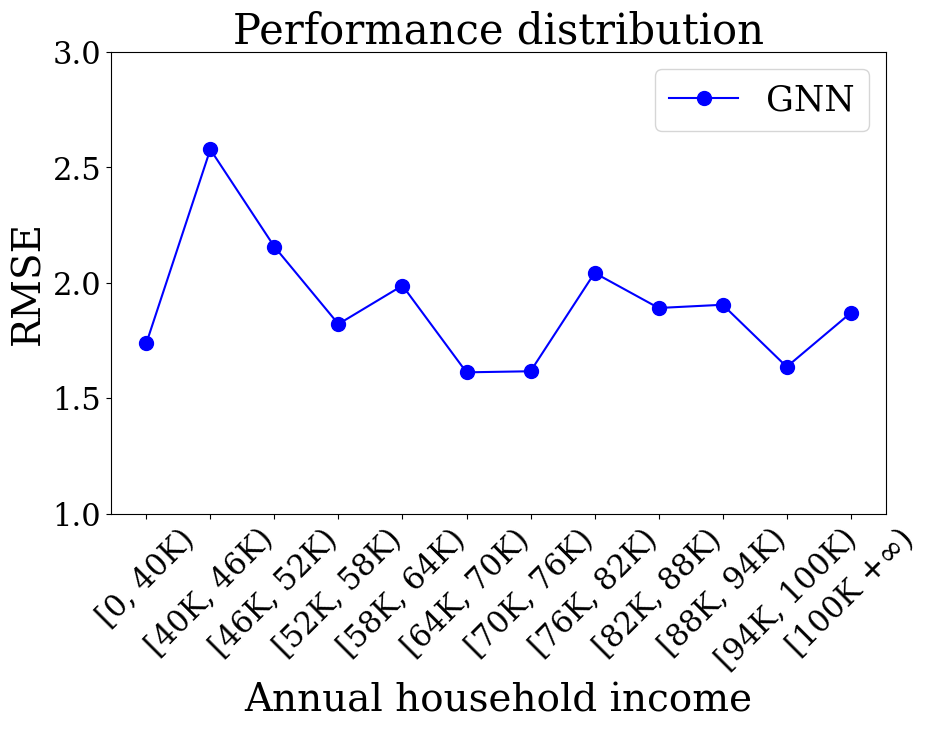}
}\hspace{-.1in}
\subfigure[]{ \label{fig:mot2}{}
\includegraphics[width=0.475\columnwidth]{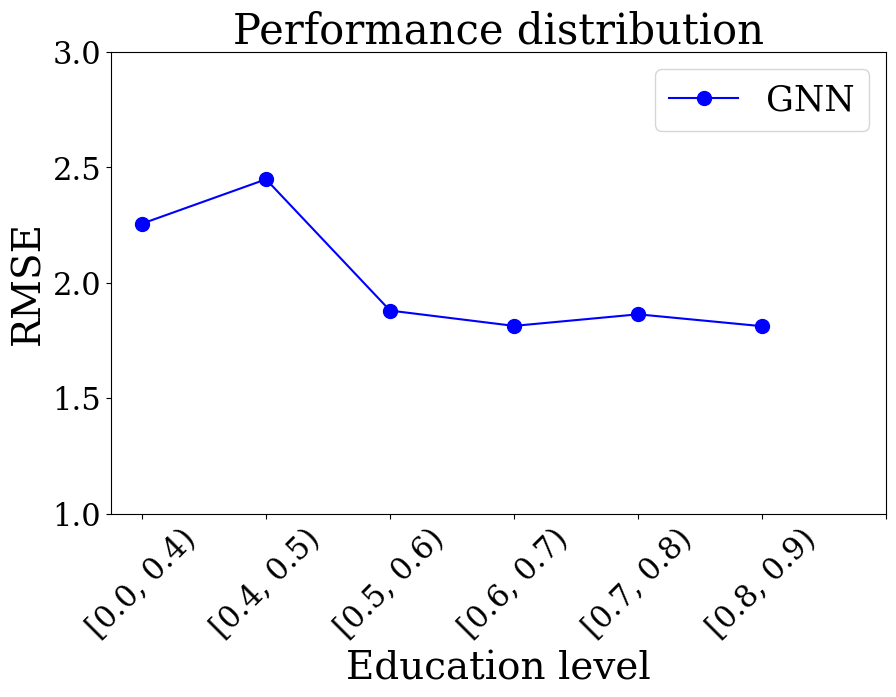}
}
\vspace{-.2in}
\caption{The distribution of RMSE for a GNN model's predictions in different sensitive values, (a) annual household income in USD, (b) education level as the proportion of the population that has attended college relative to the total population, in our study region in the Delaware River Basin.}
\label{fig:mot}
\vspace{-.2in}
\end{figure}

% bias in GNN 
Although GNNs have shown promising results in improving the predictive accuracy in initial small-scale tests, their application to large-scale stream network modeling frequently results in substantial bias related to spatial locations. 
Such spatial  bias
can lead to harm to society by amplifying discrimination and social inequity in decision and policy making, as reported in a recent National Institute of Standards and Technology (NIST) document~\cite{schwartz2022towards}. 
For example, Fig.~\ref{fig:mot} shows the distribution of temperature prediction errors (in root-mean-square errors (RMSE)) by a GNN model over locations (i.e., stream segments) with different sensitive attribute values in the Delaware River Basin. In this example, we consider two attributes, annual household income and education level. %) obtained by a GNN model for real-world water prediction in the Delaware River Basin. 
Here the education level is quantified numerically as the proportion of the population that has attended college relative to the total population within the area. It can be seen that the model has relatively poor performance in low-income and low-education communities. Such biases may unintentionally induce unfair distribution of social resources (e.g., subsidies and assistance) and treatment with respect to environmental policies, especially for low-income and low-education regions.  %where no gauging stations have been built or fewer field studies have been conducted to collect high-quality data. 
%Hence, such disparities in model performance ensure fairness in decision-making and policy development and better support regions with limited resources. %, as it substantially affects strategic decisions in water resource management and ecological policy-making, particularly in enhancing support for regions with limited data. 

% \textcolor{red}{spatial fairness, no graph, no social groups. }
Prior works~\cite{xie2022fairness,he2022sailing,he2023physics} have developed fairness formulations for spatial datasets by prioritizing training over locations with suboptimal performance, but these are not based on GNNs and fail to consider sensitive groups of different locations. 
% Researchers have also explored various approaches to promote fairness specifically for GNNs~\cite{dai2021say,bose2019compositional,kang2020inform,tang2020investigating}. Most approaches add fairness as an additional objective in the learning process. While they can mitigate the data-incurred bias (e.g., variation of data quality and data quantity), 
% they do not explicitly address the bias introduced by the model operations in GNNs. 
Recent studies have also started investigating 
the bias of GNNs based on the connections amongst different nodes
~\cite{tang2020investigating, cong2023fairsample, dong2022edits}. 
In particular, GNNs learn node embeddings by aggregating and transforming node feature vectors over the graph topology. It is possible that 
a node aggregates information from neighbors that are mostly sampled from only a few sensitive groups (e.g., high-income regions).
Therefore, the embeddings learned on the graph tend to be correlated with sensitive attributes and lead to discriminatory predictions. 
Most existing methods of addressing GNN-incurred model bias aim to balance the contribution of nodes from different sensitive groups  (e.g., high-income and low-income regions). However, they 
do not explicitly consider the influence amongst different nodes for quantifying the node contribution in the graph aggregation process. This is especially important and challenging in scientific modeling tasks due to the strong variability in node interactions driven by complex physical processes.

In this paper, we propose a Physics-Guided Fair Graph (PGFG) model to enhance fairness by leveraging the physical knowledge about node interactions. In particular, we define the influence amongst nodes in GNNs and then quantify such influence based on known physical theories about water dynamics and advection. We also utilize existing physics-based models to simulate variables that are not available or not measurable but required in computing the influence. 
Once we pre-compute the influence over the graph based on underlying physics, we introduce an edge modifier to adjust the edge sampling in the graph, with the aim to balance the influence each node receives from different sensitive groups (e.g., locations with different income levels). The edge modifier is also designed to directly balance the influence density when considering continuous sensitive attributes.

% GNN's ability to make physically consistent predictions by embedding hydrological principles into the modeling process. The latter introduces a mechanism to adjust the graph's structure, aiming to equalize the influence of nodes across different socioeconomic backgrounds, thus addressing fairness in predictions. This dual approach not only aims to improve the prediction accuracy of water temperatures but also ensures that the benefits of accurate predictions are equitably distributed. 

We evaluate our method in predicting stream water temperature using data from the past 40 years in the Delaware River Basin, which covers large areas in New York, Pennsylvania, New Jersey, Delaware, and Maryland. 
Our experiments demonstrate the superiority of the proposed method over a diverse set of baselines in preserving fairness 
over two different sensitive attributes, annual household income and educational level,  selected from the U.S. Census data~\cite{manson2020ipums}, 
without compromising predictive performance.  
%on a large-scale heterogeneous river basin, the Delaware River Basin. 

\section{Related work}

Extensive research has focused on fairness regarding predefined categorical attributes (e.g., race, gender), including regularization \cite{zafar2017fairness, yan2019fairst, kamishima2011fairness, serna2020sensitiveloss}, sensitive category de-correlation \cite{sweeney2020reducing, zhang2021towards, alasadi2019toward}, data collection or filtering strategies \cite{jo2020lessons, yang2020towards, steed2021image}, and more (e.g., a survey \cite{mehrabi2021survey}). 
Recent developments have targeted fairness formulations for spatial locations~\cite{xie2022fairness,he2022sailing,he2023physics} to promote unbiased decision-making in natural resource management, which is critical for ensuring environmental justice. 
Most of these methods aim to reduce the performance disparity (e.g., variance of performance) across locations, but neglect the relationship between bias and sensitive attributes (e.g., income, education levels, business types) over different locations. Moreover, they focus on addressing data-incurred bias (e.g., data heterogeneity and imbalance of training data), but are not designed for addressing model-incurred bias. 

GNNs have been widely used in many scientific disciplines, including hydrology~\cite{moshe2020hydronets}, freshwater science~\cite{jia2021physics_sdm}, agriculture~\cite{fan2022gnn}, and climate science~\cite{cachay2020graph}. 
Recent studies have highlighted that the learning process of GNNs may inadvertently create societal biases~\cite{dong2022edits,cong2023fairsample,he2024fair}. The inherent feature aggregation mechanisms in GNNs pose risks of propagating these biases, which affects model predictions. Traditional fairness approaches designed for independently and identically distributed (IID) data are not directly applicable to GNNs due to their inability to address biases embedded in graph structures~\cite{li2021dyadic}. 
To enhance fairness, 
%common approaches include incorporating fairness-focused regularizers into the objective function~\cite{kamishima2011fairness}. However, these methods trigger a trade-off between predictive performance and fairness. 
an approach is to modify the input graph to reduce the graph-embedded bias, typically by adjusting the adjacency matrix~\cite{li2021dyadic,laclau2021all,dong2022edits}. Research along this direction aims to create balanced node representation from diverse sensitive groups, striving for equitable embeddings. However, these methods primarily focus on balancing the number of sampled edges or adjacency weights amongst different groups, and may not adequately capture node interaction dynamics in complex scientific modeling tasks over a large space, potentially leading to aggregation bias.
% When applied to complex scientific modeling tasks over a large space, these methods do not sufficiently capture the dynamics of node interactions, which can affect the aggregation bias. Also, they mostly optimize the fairness across discrete sensitive groups but cannot handle continuous sensitive attributes.

Prior research has highlighted the importance of incorporating physical knowledge to guide machine learning in scientific modeling tasks~\cite{willard2022integrating}. Most works on this topic modify the loss function~\cite{raissi2017physics1,jia2019physics} or model structures~\cite{muralidhar2020phynet,liu2024knowledge} based on domain-specific physical knowledge. Some works also explored integrating physical relationships with the information propagation process in GNNs~\cite{jia2021physics_sdm}, which can improve the model predictive performance and generalizability even when the model is trained with limited observations. 
Incorporating physics could also provide opportunities for mitigating spatial biases, which are often caused by inadequate modeling of many complex system behaviors (e.g., dynamics and interactions). However, it remains largely under-explored how to effectively leverage such physical knowledge to promote fairness.

\section{Problem Formulation and Preliminaries}\label{sec:pd}

\noindent\underline{\textbf{Problem definition: }} The objective of this study is to predict water temperature in an interconnected river network.  We consider ${N}$ stream segments, each indexed by $i$. Each stream segment $i$ is characterized  by a set of input features over time, which is represented as   %$\textbf{X}_i$, represented as 
$\textbf{x}_i = \{\textbf{x}_{i,1}, \textbf{x}_{i,2}, ..., \textbf{x}_{i,T}\}$, collected over $T$ time steps (e.g., dates). 
These features include meteorological variables like daily average solar radiation, air temperature, precipitation, local wind speed, date of the year, and some modeled local variables (refer to details in the technical appendix). 
Additionally, observed target variables $\{{y}_i^t\}$ (i.e., water temperature) are available for certain segments $i \in \{1, ..., N\}$ at certain time steps $t \in \{1, ..., T\}$. 

This study also focuses on promoting fairness in model performance. Fairness is measured with respect to a specific sensitive census attribute $\textbf{S}$ (e.g., income level), which can be either categorical or continuous. This attribute is associated with regions (e.g., subcounties) through which each stream segment flows. 
For each stream segment $i$, we aggregate the sensitive attribute values from all overlapping and adjacent regions, and we represent this aggregated sensitive attribute value as $s_i$.
Our objective is to ensure that the model's predictive accuracy is equitable over locations with different attribute values (e.g., low-income and high-income regions).

\noindent\underline{\textbf{Graph representation of stream networks:}}
Graphs are widely used for representing the interactions amongst stream segments in a stream network~\cite{jia2021physics_sdm,chen2022physics}. Here a graph is defined as $G=\{\mathcal{V},\mathcal{E}\}$, where the node set $\mathcal{V}$ represent stream segments, and the edge set $\mathcal{E}$ represent connections between upstream-downstream segment pairs. We create an edge between node $i$ and node $j$ if node $i$ is anywhere upstream from node $j$.
The edges in the graph can be assigned weights
based on the inverse of the stream distance between two stream segments (refer to details in the dataset section). These edge weights are stored in a node adjacency matrix $\textbf{A}$. GNNs leverage this graph structure for predictive analysis with two key benefits: (1) They aggregate contextual information from connected segments to enhance the accuracy in predicting target variables, which are often highly influenced by water dynamics in upstream segments. (2) GNNs enable the information diffusion of observational data from data-rich to data-scarce stream segments, which helps improve the overall performance.

\noindent\underline{\textbf{Fairness formulation: }}
A variety of fairness formulations have been proposed, broadly categorized into three types: (i) individual fairness~\cite{lahoti2019ifair,lahoti2019operationalizing} that treats similar individuals similarly, (ii) group fairness~\cite{hardt2016equality,dai2021say} that expects parity of statistical performance across groups, and (iii) fairness formulations that aim to improve per-group performance, such as Pareto-fairness~\cite{ge2022toward} and Rawlsian Max-Min fairness~\cite{lahoti2020fairness,zhang2014fairness}. In this work, we extend the formulations (ii) and (iii) to the spatial data of a large number of stream segments with respect to a specific continuous sensitive attribute (e.g., annual household income).

To evaluate fairness in the context of a continuous sensitive attribute, one approach would be to first discretize the sensitive values into distinct groups, e.g., different income intervals. This allows us to apply group fairness concepts effectively. Here we introduce the fairness measure $M_{fair}$, defined on a set of groups $\p$. This metric is designed to balance model performance across all groups within $\p$, as represented by the following equation:
\begin{equation}\label{eq:e}
M_{fair}(\mf_\vtheta, M_\mf, \p) =
\sum_{p\in\p}\frac{d(M_\mf(\mf_\vtheta, p), E_{\p})}{|\p|},
\end{equation}
where $M_\mf$ represents a performance metric (e.g., RMSE) for model $\mf_\vtheta$ on a specific group $p$, $E_\p$ denotes the average model performance across groups, and $d(\cdot, \cdot)$ is a distance function. Intuitively, a significant deviation of a group's performance $M_\mf(\mf_\vtheta, p)$ from the overall average $E_\p$ indicates potential unfairness in model $\mf_\vtheta$. In our test, we calculate $E_\p$ as the overall performance of the model $\mf_{\vtheta}$ across all groups, which is represented as $E_\p = M_\mf(\mf_{\vtheta}, \{\cup p\,| \,  p\in\p\})$; and $d(\cdot, \cdot)$ is measured as the absolute distance.

% \begin{equation}\label{eq:e}
%     % E_\p = \sum_{p\in\p} \frac{N_p \cdot M_\mf(\mf_{\vtheta_0}, p)}{N}
%     E_\p = M_\mf(\mf_{\vtheta_0}, 
%     \{\cup p\,| \,  p\in\p\})
%     % E_\p = \sum_{p\in\p} \frac{M_\mf(\mf_{\vtheta_0}, p)}{|\p|}
% \end{equation}

% \textcolor{red}{evaluation}
In this work, we also propose new algorithms to directly improve the fairness with respect to the continuous attributes. In the evaluation phase, we assess the model's effectiveness in reducing spatial bias by measuring its performance in worst-case scenarios. 
Ideally, the model should perform consistently across all ranges of the sensitive attribute values.
We use a sliding window with a predetermined size %. As the window traverses 
to traverse the entire range of attribute values. As it moves, we evaluate the model's predictive performance for stream segments within the current window. Our fairness objective is to improve the model performance in the worst-case scenarios identified by any window. %model performance in any window. %The primary objective of fairness is to identify and enhance performance in the worst-case scenario.

\section{Proposed Method}

%\textcolor{red}{Add overview. }
% The visual overview of the proposed method is shown in Fig.~\ref{fig:overall_flow}. We begin by introducing the model architecture. Then we delve into the integration of physical knowledge, which effectively estimates the influence between neighboring nodes. Finally, we explore the edge sampling approach by leveraging the quantified physical knowledge to enhance the fairness.

\begin{figure*}[h]
	\centering
\subfigure[]{ \label{fig:flow1}{}
\includegraphics[width=1.0\columnwidth]{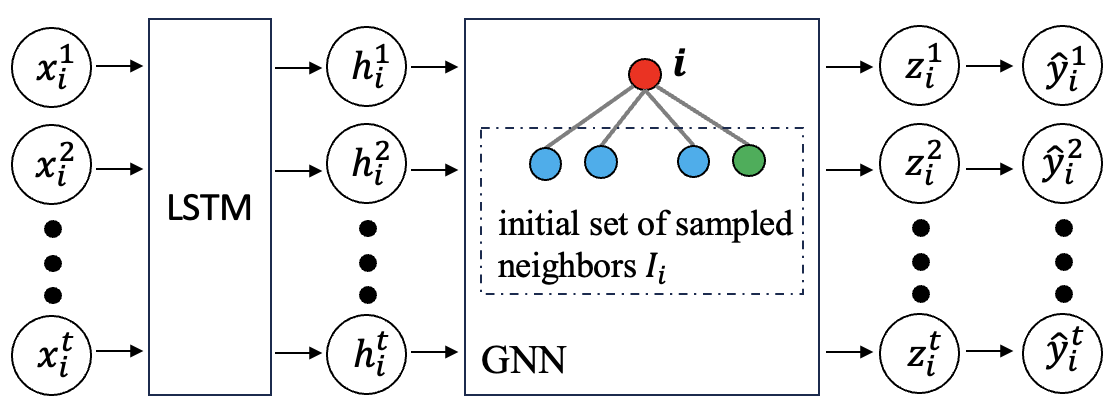}
}
\hspace{0.2cm}
\subfigure[]{ \label{fig:flow2}{}
\includegraphics[width=1.0\columnwidth]{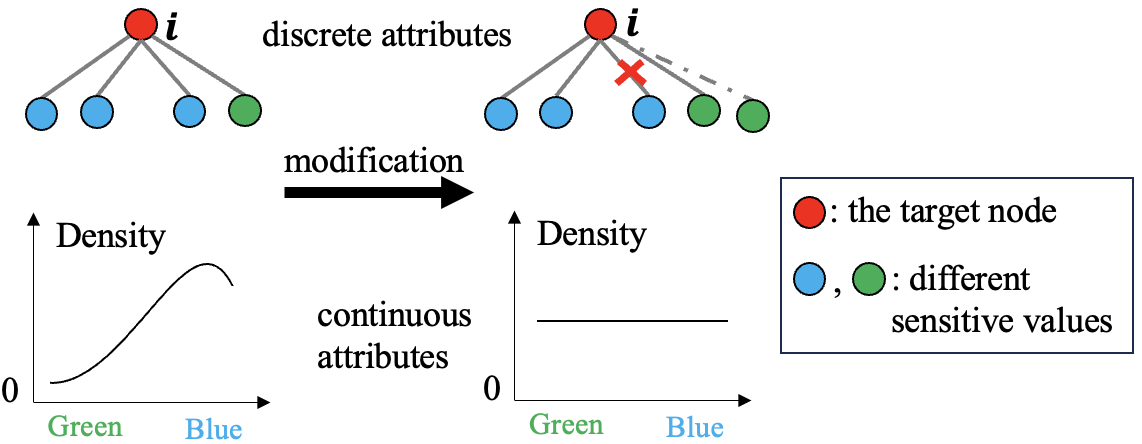}
}
\vspace{-.2in}
	\caption{(a) A diagram of the proposed model. For each node $i$ and each time $t$, the long short-term memory (LSTM) network extracts an embedding $\textbf{h}_{i,t}$. Then we apply a graph neural network (GNN) to refine each time's embedding by aggregating information from neighboring nodes (highlighted in blue and green), producing a new embedding $\textbf{z}_{i,t}$. Finally, the fully connected layers output the prediction $\hat{\textbf{y}}_i^t$. (b) Fair edge sampling for discrete and continuous sensitive attributes.}
	\label{fig:overall_flow}
\vspace{-.1in}
\end{figure*}

\subsection{Model Architecture}

We develop a machine learning model architecture $\mathcal{F}$ to simulate water temperature in stream segments by taking into account their spatial and temporal dependencies (Fig.~\ref{fig:overall_flow}). For each stream segment, its thermal status and water quantity change gradually based on the current weather input and its historical state. We use a long short-term memory (LSTM) network layer to capture this temporal dependency. This layer generates the hidden presentation $\{\textbf{h}_{i,t}\} = \text{LSTM}(\{\textbf{x}_{i,t}\})$ for each segment $i$ at each time step $t$ by integrating the current input $\textbf{x}_{i,t}$ with the previous LSTM state and hidden representation $\textbf{h}_{i,t-1}$.

The water temperature in a stream segment is also affected by its neighbors, e.g., through water advected from upstream segments. % and thermal exchange with downstream segments. 
Hence, after gathering the hidden representations for all the segments, we use $L$ graph convolutional layers to capture interactions between neighboring stream segments.
Formally,
the $l$-th layer of graph convolution
can be expressed as follows: 
\begin{equation}
    \begin{aligned}
    \textbf{z}_{i, t}^{(l)} &= g_a^{(l)}([\textbf{z}_{i, t}^{(l-1)}, \textbf{a}_{i, t}^{(l)}];\theta_a^{(l)}), \text{for}\,\,\, l \in \{1, ..., L\}, \\
    \textbf{a}_{i, t}^{(l)} &= \text{Pooling}(\{\textbf{z}_{j, t}^{(l-1)}, \forall j \in \mathcal{N}(i)\}, \textbf{A}_{i.}), 
    \end{aligned}
    \label{eq:aggr}
\end{equation}
where $\textbf{z}_{i, t}^{(0)}=\textbf{h}_{i,t}$, 
$\mathcal{N}(i)$ denotes the set of neighboring stream segments to $i$, and $\theta_a^{(l)}$ represents the parameters in the $l$-th graph convolution layer. The latent representation $\textbf{a}_{i, t}^{(l)}$ embeds the information from neighboring stream segments, and is obtained through a weighted pooling of embeddings from the neighbors based on the weights from the adjacency matrix $\textbf{A}$. We concatenate $\textbf{a}_{i, t}^{(l)}$ with the last GNN layer's embedding of the target node $\textbf{z}_{i, t}^{(l-1)}$ before the transformation using the function $g_a(\cdot)$. In this work, we adopt the GraphSAGE method~\cite{hamilton2017inductive} to implement the function $g_a(\cdot)$. Finally, a series of fully connected output layers are stacked to transform the aggregated embeddings $\textbf{z}_{i,t}$ into the predicted output $\hat{{y}}_{i,t}$.

The model is updated by minimizing the distance between the predictions and true observations, as follows:
\begin{equation}
\mathcal{L} = \sum_{{i,t}\in\mathcal{D}} ||\hat{y}_{i,t}-y_{i,t} ||^2/|\mathcal{D}|,
\label{eq:mse}
\end{equation}
where $\mathcal{D}$ is the set of $\{i,t\}$ for which the observed $y_{i,t}$ is available. 

\subsection{Physical Knowledge Integration}
\label{sec:phy}
Despite the promise of GNN in capturing spatial dependencies, the aggregation process (Eq.~\ref{eq:aggr}) can introduce learning bias over different groups. When training the GNN model, the prediction errors (Eq.~\ref{eq:mse}) on each node $i$ will be propagated to both node $i$ and its neighbors because the hidden representation $\textbf{z}_j$ for any node $j\in \mathcal{N}(i)$ is involved in computing the latent representation $\textbf{a}_i$ for node $i$. 
Hence, the errors from the node $i$ could also contribute to the training of its neighboring nodes. 
However, this aggregation process could lead to a group-biased learning process when the neighbors' influence is unevenly distributed over different groups.  Here we first introduce a novel physics-guided graph approach to quantify the influence between neighboring nodes. The proposed influence measure will then be used to create a new fairness-driven aggregation method, which will be discussed later in the next section.

Based on the graph aggregation process, for each node $i$, we define the influence it receives  from its neighbor node $j$ at time $t$ as follows:
\begin{equation}
\text{Influence}_t(j,i) = ||\frac{\partial \hat{y}_{i,t}}{\partial \textbf{z}_{j,t}}||, \,\,\text{for}\,\, j\in \mathcal{N}(i), 
\label{eq:influence}
\end{equation}
where $||\cdot||$ represents the L2 norm of a vector. 

We could estimate the gradients over the GNNs to compute the influence  
during the training process. However, this may degrade the training performance and stability because the influence estimate is dependent on aggregation parameters $\theta_a$ and thus can be affected by an immature model. The updated aggregation using the influence estimate will also in turn affect the learning of aggregation parameters $\theta_a$. 
More importantly, true observations $\{y_{i,t}\}$ are often sparse over space and time in scientific modeling tasks, e.g., some low-income or remote regions have limited budgets in collecting high-quality observations, which leads to initial bias in the gradient estimates and $\theta_a$.
Moreover, directly utilizing these gradient estimates requires additional computational costs.

To address these challenges, we propose to leverage underlying physical knowledge to estimate the influence over the graph. Most dynamical systems are governed by partial differential equations (PDEs), which describe the change of target variables in response to external drivers and spatial interactions. When modeling a set of locations as different nodes in a graph, we can often estimate the influence between a pair of nodes based on the spatial interaction terms (represented as spatial derivatives) in the governing PDE. In stream networks, the dynamics of stream water temperature are governed by the heat transfer process following the law of energy conservation~\cite{dugdale2017river}. The heat transfer process involves different heat fluxes, including shortwave and longwave radiation, latent heat flux (e.g., evaporation), and advective and convective heat flux. In particular, the advective heat flux describes the interaction between different stream segments. Through some derivations and approximations from the heat transfer PDE, we can obtain the relation of water temperature between neighboring segments~\cite{dugdale2017river,boyd1996heat}, as follows: 
\begin{equation}
y_{i,t}  \approx %\frac{q_{j,t}}{q_{i,t}+q_{j,t}} y_{j,t}, \,\, 
\frac{y_{i,t-1} q_{i,t}+ \sum_j y_{j,t} q_{j,t}}{ q_{i,t}+q_{j,t}}, 
\text{for}\,\, j\in \mathcal{N}(i), 
\label{eq:phy}
\end{equation}
where $q_{i,t}$ represents the streamflow at the stream segment $i$ at time $t$. 
Then we can obtain $\frac{\partial y_{i,t}}{\partial y_{j,t}}$ as  
\begin{equation}
\frac{\partial y_{i,t}}{\partial y_{j,t}} = \frac{q_{j,t}}{q_{j,t}+q_{i,t}}, \text{for}\,\, j\in \mathcal{N}(i)
\label{eq:partial}
\end{equation}

Finally, we can approximate the influence from node $j$ to node $i$ as follows: 
\begin{equation}
\begin{aligned}
\text{Influence}_t(j,i) &\approx ||\frac{\partial y_{i,t}}{\partial y_{j,t}} \frac{\partial \hat{y}_{j,t}}{\partial \textbf{z}_{j,t}} ||\\
&=\frac{q_{j,t}}{q_{j,t}+q_{i,t}} ||\frac{\partial \hat{y}_{j,t}}{\partial \textbf{z}_{j,t}} ||, \,\,
\text{for}\,\, j\in \mathcal{N}(i)
\end{aligned}
\end{equation}

When we compare the influence between different pairs of nodes,  the term $||\frac{\partial y_{j,t}}{\partial \textbf{z}_{j,t}} ||$ is constant because 
all the nodes use the same transformation layers to convert $\textbf{z}$ to the output $\hat{y}$. Hence we  just need to consider $\frac{q_{j,t}}{q_{j,t}+q_{i,t}}$ for the influence from node $j$ to node $i$.  

As the streamflow observations $q$ are very sparse for most stream segments, we  complement observed streamflow with physical simulations. We leverage a physics-based PRMS-SNTemp model~\cite{markstrom2012p2s} to simulate streamflow for all the stream segments and all the time steps. We use the simulated value when the streamflow $q_{i,t}$ is not available for certain segments or certain dates. Then we can use such combined streamflow data to estimate the influence between each pair of neighboring nodes.

Another issue is that the physical relationship described by Eq.~\ref{eq:phy} only applies to a node $i$ and its direct upstream segment $j$. Most stream segments only have a few number of upstream segments, which makes the size of  neighborhood relatively small. To enhance the power of GNN in leveraging spatial dependencies, prior work in this domain often extends the neighborhood to multi-hops away~\cite{jia2021physics_sdm,chen2022physics}, i.e., edges are created between node $j$ and node $i$ as long as node $j$ is anywhere upstream from node $i$. This often leads to better performance because each node $i$ can still be affected by water flow from a distant upstream segment $j$. %the water flow advected from node $j$ can still affect node $i$ even though they are distant away. 
When segment $j$ to segment $i$  are not geographically adjacent, we estimate the influence from node $j$ to node $i$ following the stream path between $j$ and $i$, which is represented as \{$j_0=j$, $j_1$, $j_2$, ..., $j_M=i$ \}. The influence can be estimated as follows: 
\begin{equation}
\text{Influence}_t(j,i) = \prod_{m=1}^M \frac{q_{j_{m-1},t}}{q_{j_{m-1},t}+q_{j_m,t}}
\end{equation}

This approach can be efficiently implemented as the influence over the graph can be pre-computed using the streamflow data. 
This approach is also generally applicable to many other scientific modeling tasks (e.g., hydrology, climate simulation, molecule modeling in material science) because we can often derive the spatial dependencies of target variables $\frac{\partial y_i}{\partial y_j}$ from the governing PDEs.

% \textcolor{red}{generalizabiltiy to other domains. }

\subsection{Fairness-driven Edge Sampling}
\label{sec:sampling}

Prior works on mitigating bias in GNNs ~\cite{li2021dyadic,cong2023fairsample, laclau2021all,dong2022edits} are focused on balancing the number of edges or the sum of adjacency weights across different sensitive groups, but they do not sufficiently consider the node influence and how that will bring bias to the learning process.

We introduce an edge modifier that leverages physics-based influence estimates to 
update the edge connections amongst nodes so as 
to improve group fairness. 
For each node $i$, we randomly sample a fixed number of initial neighbors following the GraphSAGE ~\cite{hamilton2017inductive}. The edge modifier then either injects or removes edges to balance the physical influence across different sensitive groups within the sampled neighborhood of node $i$. 
Within the candidate neighbors for each sensitive group, we prioritize selecting neighbors that have similar features $\textbf{x}$ with the target node (i.e., the locations under similar weather conditions).  This  help mitigate noise in $\textbf{x}$ gathered through aggregation and improve the model robustness. 
% In addition, accuracy in node prediction can be enhanced by aggregating embeddings from informative neighbors. A neighbor node $u$ is considered informative for a node $v$ if their feature vectors are similar. 
% %Based on this intuition, our edge modifier specifically proceeds as follows. 
% However, as we only pick informative neighbors for GNN training, the noisy edges may not be used and thus their negative effect on the accuracy of the GCN may be reduced. 

After these edge modification steps, we slightly rescale the edge weights to balance the influence from different groups. For each node $i$, we first measure the sum of influence from all the neighbors in each sensitive group $\mathcal{G}_k$ %of neighbors 
as 
$\text{SI}_k = \sum_{j\in \mathcal{G}_k} \text{Influence}_t(j,i)$. Then we rescale the weight of each edge $(j,i)$ such that  the influence from each group matches the highest group influence,   as follows:
\begin{equation}
\textbf{A}'_{ji} = \textbf{A}_{ji}\frac{\max_{k'} \text{SI}_{k'}}{\text{SI}_k}, \,\, \text{for}\,\, j\in \mathcal{G}_k. 
\end{equation}

% This approach ensures we sample equal number of neighbors for each node. Also, in our implementation, all the neighbors and weights can be 

% the influence to efficiently balance equal influence across different groups and adhere to the GraphSAGE principle of limiting the number of sampled neighboring nodes, we assign aggregation weights to different groups to balance influence. This reduces the computational cost and incurs only a small extra computational overhead for the edge sampling process, making it more computationally efficient than the majority of the existing graph modification methods.

To train fair GNNs with continuous sensitive attributes, the edge modifier aims to generate a more balanced density distribution of influence over sensitive attribute values from each node's neighborhood. %  neighboring nodes for a target node. 
The central idea is to modify the edge connections to maximize the variance of sensitive values. Similar to the case with discrete attributes, we first sample an initial set of neighbors $\mathcal{I}_i$ for each node $i$, and represent other candidate neighbors as $\mathcal{C}(i)$. 
%%%
Then for each of other candidate neighbors $j\in$ $\mathcal{C}(i)$, we quantify the density at $j$ by aggregating the influence of initial neighbors based on their similarity to the candidate neighbor, as follows: %similarity of sensitive attribute values between each candidate neighbor $j$ over the initial neighbors in $\mathcal{I}_i$. % striving to maximize this aggregate sum. 
%The formula is as follows:
\begin{equation}
\begin{aligned}
    \text{Density}(j) &= \sum_{j^\prime} \text{sim}(s_j,  s_{j^\prime})\cdot \text{Influence}(j^\prime, i), \\
    &\text{for}\,\,j\in \mathcal{C}(i), j^\prime\in \mathcal{I}_i,
\end{aligned}
\end{equation}
where $\text{sim}(a,b)$ is the similarity function and measured as $1- Norm(|a-b|)$, which normalizes and inverts to ensure lower $|a-b|$ indicate greater similarity.
%as $1/(1+|a-b|)$.  

We prioritize adding node $j$ as neighbors and including it in the aggregation process if it has a lower density value. % for node $j$ indicates a preference to include it in the aggregation process, facilitated by injecting an edge between $j$ and $i$. 
Intuitively, if a candidate node $j$ has a sensitive attribute value highly similar to many existing neighbors and these neighbors have already exerted significant influence, 
we opt not to include node $j$ as a neighbor to avoid amplifying biases in the aggregation phase. 
Instead, we prefer to include candidate nodes that have their sensitive attribute values different from existing highly influential neighbors.  %with lower influence or lower similar sensitivity to balance the high-influence nodes. 
%Finally, to ensure a uniform distribution of influence, we assign aggregation weights to each sampled node. Since the number of sampled nodes for the aggregation process is fixed, we also compute scores for these sampled nodes and remove edges for those with higher scores, thereby refining the edge sampling process. 
%In our test, the $sim(\cdot, \cdot)$ is the Euclidean distance as the sensitive value is a vector.

\section{Experiments}
%In this section, we conduct experiments on a large-scale heterogeneous river basin to evaluate the effectiveness of PGFG.

\subsection{Dataset}
\label{sec:dataset}
% The dataset we use is collected from the Delaware River Basin, which is an ecologically diverse region 
% and a watershed along the east coast of the United States that provides drinking water to over 15 million people~\cite{williamson2015summary}.
% %a societally important watershed along the east coast of the United States as it provides drinking water to over 15 million people~\cite{williamson2015summary}.
The dataset used in our evaluation is collected from the Delaware River Basin (DRB), sourced from the U.S. Geological Survey's National Water Information System~\cite{us2016national} and the Water Quality Portal~\cite{read2017water}. 
Observations at specific latitudes and longitudes are matched to stream segments that vary in length from 48 to 23,120 m. These segments are defined by the geospatial fabric used for the National Hydrologic Model~\cite{regan2018description}, and are split up to have roughly a 1-day water travel time. We match observations to stream segments by snapping observations to the nearest stream segment within a tolerance of 250 m. Observations farther than 5,000 m along the river channel to the outlet of a segment are omitted from our dataset. Segments with multiple observation sites are aggregated to a single mean daily water temperature value. 
Refer to ~\cite{sam_release} for the full observational dataset. More details of the dataset are in the technical appendix.

DRB contains 456 stream segments with input features at the daily scale from Jan 01, 1980, to Jul 31, 2020 (14,823 dates). 
% The input features include meteorological variables like daily average solar radiation, air temperature, precipitation, local wind speed, date of the year, solar radiation, shade fraction, potential evapotranspiration, and the geometric features of each segment (e.g., elevation, length, slope, and width) ~\cite{sam_release}. 
% Air temperature and precipitation values were derived from the GridMET gridded dataset~\cite{gridMET}. Other input features (e.g., shade fraction, solar radiation, potential evapotranspiration) are difficult to measure frequently, and we use values produced by the PRMS-SNTemp model~\cite{markstrom2012p2s} as its internal variables.
In addition, amongst 456 segments, the number of streamflow observations only available for the 183 segments ranges from 16 to 14,774 with a total of 1,928,445 streamflow observations across all dates and segments. For those segments or specific dates where streamflow data are not available, we utilize simulated data produced by the physics-based PRMS-SNTemp model~\cite{markstrom2012p2s}.

% \textcolor{red}{add streamflow and simulated data for physical knowledge}
 
% \textcolor{red}{add information for census information.}

\begin{table*}[]
    \scriptsize
    \centering
    \caption{The fairness and overall RMSE with two continuous sensitive attributes. The lower value indicates the better result. Bold indicates the best-performing model for a given metric and attribute, and '$-$' indicates that the model is not applicable.}
    \vspace{-.1in}
    \begin{tabular}{|c|cc|cccc|cc|cccc|}
    \hline
    \multirow{3}{*}{Method} & \multicolumn{6}{c|}{Annual household income} & \multicolumn{6}{c|}{Education level} \\ \cline{2-13} 
    & \multicolumn{2}{c|}{Discrete scenario} & \multicolumn{4}{c|}{Continuous scenario} & \multicolumn{2}{c|}{Discrete scenario} & \multicolumn{4}{c|}{Continuous scenario} \\ \cline{2-13} 
    & RMSE & \multicolumn{1}{c|}{Fairness} & RMSE & 1000 & 3000 & 5000 & RMSE & \multicolumn{1}{c|}{Fairness} & RMSE & 0.003 & 0.006 & 0.010 \\ \hline\hline
    GraphSAGE & 1.774 & 0.206 & 1.774 & 5.057 & 4.384 & 3.268 & 1.774 & 0.253 & 1.774 & 4.910 & 4.331 & 3.959 \\ \hline
    REG & 1.785 & 0.181 & 1.765 & 4.802 & 4.040 & 3.073 & 1.786 & 0.226 & 1.765 & 4.625 & 4.323 & 4.007 \\ \hline
    FairGNN & 1.774 & 0.157 & 1.784 & 4.864 & 3.807 & 3.022 & 1.788 & 0.215 & 1.767 & 4.477 & 4.201 & 3.760 \\ \hline
    Bi-Level & 1.776 & 0.183 & 1.822 & 4.778 & 3.555 & 2.908 & 1.790 & 0.206 & 1.822 & 4.490 & 4.317 & 3.815 \\ \hline
    DSGNN$_g$ & 1.832 & 0.189 & $-$ & $-$ & $-$ & $-$ & 1.829 & 0.225 & $-$ & $-$ & $-$ & $-$ \\
    DSGNN & 1.850 & 0.141 & 1.850 & 5.025 & 3.474 & 3.026 & 1.850 & 0.210 & 1.850 & 4.684 & 4.311 & 3.720 \\ \hline
    SLGSGNN$_g$ & 1.794 & 0.175 & $-$ & $-$ & $-$ & $-$ & 1.792 & 0.227 & $-$ & $-$ & $-$ & $-$ \\
    SLDSGNN & 1.791 & 0.185 & 1.791 & 4.679 & 3.453 & 2.767 & 1.791 & 0.240 & 1.791 & 4.252 & 3.988 & 3.988 \\ \hline \hline
    FairEdge & 1.783 & 0.196 & 1.780 & 4.751 & 3.975 & 2.901 & \textbf{1.773} & 0.220 & 1.774 & 4.478 & 3.965 & 3.846 \\ \hline
    FairAdj & \textbf{1.767} & 0.170 & 1.784 & 4.709 & 3.671 & 2.715 & 1.780 & 0.206 & \textbf{1.759} & 4.219 & 3.991 & 3.654 \\ \hline
    PGFG & 1.774 & \textbf{0.134} & \textbf{1.748} & \textbf{4.456} & \textbf{3.333} & \textbf{2.573} & 1.783 & \textbf{0.189} & 1.777 & \textbf{3.999} & \textbf{3.817} & \textbf{3.597} \\ \hline
    \end{tabular}
\label{tab:performance}
\end{table*}

We take sensitive attributes from the U.S. Census data~\cite{manson2020ipums} and process them at the subcounty level. Here we focus on two sensitive attributes,  median annual household income and education level. These attributes are considered in the context of two types of fairness: group fairness and fairness with continuous sensitive attributes. For median annual household income, we categorize it into three distinct groups: low income (0-50,000), middle income (50,000-100,000), and high income ($>$100,000). In this categorization, we identify 43 streams in the low-income group, 329 streams in the middle-income group, and 84 streams in the high-income group. Education level is quantified numerically based on the proportion of the population that has attended college relative to the total population within each subcounty. These data are then divided into two categories: a low-education group, where the proportion is less than 0.5, and a high-education group, where the proportion exceeds 0.5. Within these education categories, 69 streams are classified within the low-education group and 387 streams within the high-education group. Finally, we assign a sensitive attribute value to each stream segment based on the average sensitive values of all the subcounties it flows through.

\subsection{Implementation details}

We generate the adjacency matrix $\textbf{A}$ based on the river distance between each pair of stream segment outlets, represented as $\text{dist}(i, j)$. We standardize the stream distance and then compute the graph edge weights as $\textbf{A}_{ij} = 1/(1+\text{exp}(\text{dist}(i,j)))$. More details are in the technical appendix.

We implement different candidate approaches with the initial adjacency matrix for 100 epochs (converged), using the Adam optimizer ($\alpha=0.001$). %We use the first 2/3 of the dataset for training and the remaining 1/3 for testing. 
In the following experiments, we use data from the first 27 years (Jan 01, 1980, to Jan 20, 2007) for training and then test in the next 13 years (Jan 21, 2007, to Jul 31, 2020). 
Within the training data, the last 1/3 of the time steps are further separated as validation data to halt training so as to prevent overfitting.

\subsection{Candidate methods}

We compare the performance and fairness of the proposed method with several representative baselines.

\begin{itemize}[leftmargin = *]
    \item GraphSAGE~\cite{hamilton2017inductive} is a widely used GNN model without fairness constraints, which only aims to optimize model predictive performance by randomly  sampling an equal number nodes for embedding aggregation. It also serves as the base model for our proposed method and all the baselines. 
    
    \item REG~\cite{serna2020sensitiveloss} is a commonly used method for enforcing fairness by including the fairness objective as a regularization term in the loss function. In our implementation, 
    for discrete attributes, 
    REG includes the fairness metric by Eq.~\ref{eq:e} as the regularization term in the loss function. For continuous attributes, the regularization term is defined as the variance of performance across nodes.
    
    \item FairGNN~\cite{dai2021say} adopts an adversarial discriminator to rectify the final layer representations in the GNN model. Given the representation $\textbf{z}$, the adversary aims to predict the node's sensitive group for discrete attributes and the specific sensitive values for continuous attributes. %or specific sensitive value for discrete attributes and continuous attributes, respectively.

    \item Bi-Level~\cite{xie2022fairness} 
    introduces a bi-level optimization that adjusts learning rates via a global referee based on predictive performance. When considering discrete attributes or continuous attributes, it assigns relatively higher learning rates for underperforming groups or nodes, enhancing fairness without compromising model accuracy. This bi-level design helps disentangle model prediction and fairness objectives.

    \item Both DSGNN and SLDSGNN utilize degree-specific parameters to improve fairness~\cite{tang2020investigating}. Specifically, SLDSGNN adds self-supervised learning for pseudo-labeling unlabeled nodes. To further adapt these models for discrete attributes, we extend these baselines by including group-specific parameters (using a similar method as in~\cite{tang2020investigating} for learning degree-specific parameters). We denote by DSGNN$_g$ and SLDSGNN$_g$ the variants of DSGNN and SLDSGNN with group-specific adjustments.

    \item PGFG is the proposed method that integrates physical knowledge and the edge modifier. 
\end{itemize}

%GraphSage~\cite{hamilton2017inductive} is a famous GNN model without fairness constraints, which only aims to optimize model predictive performance by randomly sampling nodes for embedding aggregation. REG~\cite{zafar2017fairness, yan2019fairst, kamishima2011fairness, serna2020sensitiveloss} is a main strategy used for general fairness-aware learning by enforcing fairness as a regularization term to the loss function. FairGNN~\cite{dai2021say} adopts an adversarial discriminator to rectify the final layer representations in the target GNN model. Bi-Level~\cite{xie2022fairness} proposes a bi-level model refinement to disentangle model prediction and fairness objective. This method effectively addresses the competition between predictive accuracy and fairness that exists in the previous two baselines (REG and FairGNN). Both DSGNN and SLDSGNN\cite{tang2020investigating} apply the degree-specific parameters, however, SLDSGNN uniquely integrates self-supervised learning that creates pseudo-labels on unlabeled nodes. PGFG is the proposed method that integrates physical knowledge and the edge modifier. 
We include two variants of the proposed method to show the effectiveness of incorporating physical knowledge: (1)~FairEdge, which only balances the number of edges across different sensitive values, and (2) FairAdj, which replaces the physical influence with the edge weights in the initial adjacency matrix $\textbf{A}$.

We also compare our method with two spatio-temporal GNNs, STGNN~\cite{wang2020traffic} and FairFor~\cite{he2023learning}, and a physics-guided graph network PGRGrN~\cite{jia2021physics}; see the technical appendix for details of results.

\subsection{Results}\label{sec:results}

\noindent\textbf{Overall accuracy and fairness evaluation: }
Table~\ref{tab:performance} presents a comprehensive overview of the predictive performance and fairness of our proposed method and other baselines for two selected sensitive attributes. We utilize the RMSE as the evaluation metric. This metric measures the average magnitude of the errors between predicted and observed temperatures. The fairness performance is measured using two distinct metrics, i.e., performance disparity (discrete attribute) and worst-case performance (continuous attribute),  as discussed in previous section. 
For the sensitive attribute of annual household income, we test three sliding window sizes: 1000, 3000, and 5000, and report the worst-case RMSE for each one. For the sensitive attribute of education level, we also try three sizes for the sliding window: 0.003, 0.006, and 0.01. For these metrics, a lower value indicates better performance.

Metrics in Table~\ref{tab:performance} indicate that the proposed PGFG method outperforms the baselines in terms of fairness for both discrete and continuous attribute scenarios. Moreover, PGFG effectively promotes the fairness without compromising the predictive performance. We can also observe that  FairEdge has comparable fairness performance with other explicit fairness-preserving methods (e.g., REG and FairGNN), which shows the promise in promoting fairness through the modification of neighborhood in graphs. % the neighborhood in promoting the fairness. 
PGFG performs much better than both FairEdge and FairAdj. This confirms that merely balancing the number of neighbors or the sum of weights from different groups is not sufficient; it is also necessary to consider their physical influence. %shows the effectiveness of refining the graph neighborhood in mitigating the bias during the aggregation process.

% In addition, we observe that the proposed PGFG method outperforms the FairAdj method, which shows the benefit of embedding physical knowledge. The improvement from FairAdj to FairEdge confirms the benefit of analyzing the node influence. 

\begin{figure} [!t]
\centering
\subfigure[]{ \label{fig:g_income}{}
\includegraphics[width=0.475\columnwidth]{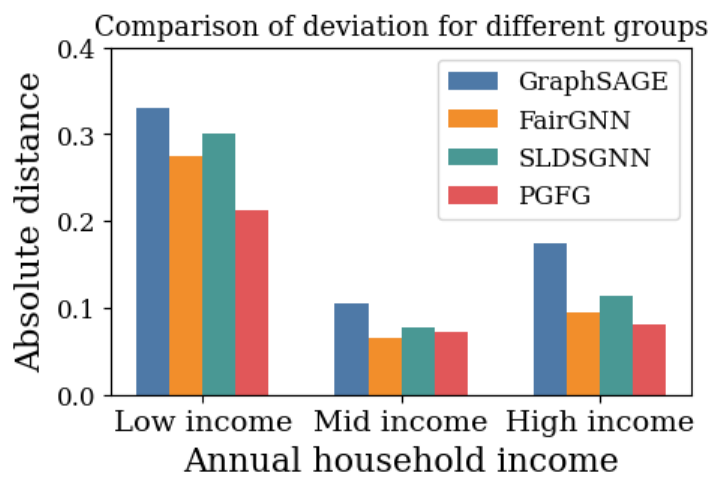}
}%\vspace{-.1in}
\subfigure[]{ \label{fig:g_edu}{}
\includegraphics[width=0.475\columnwidth]{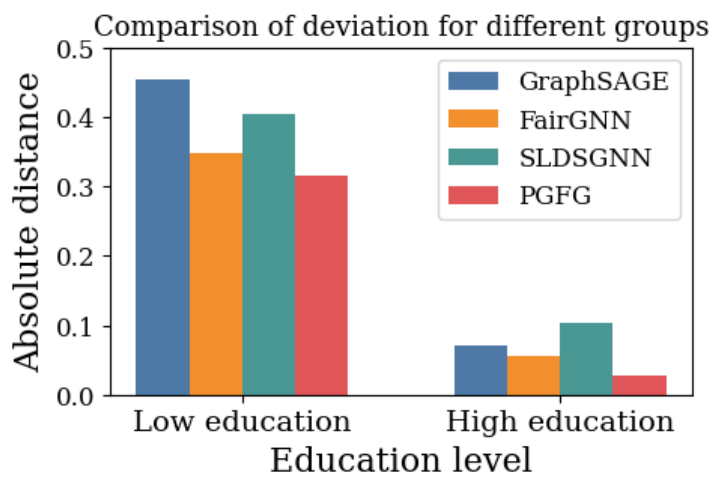}
 }
 \vspace{-.2in}
\caption{Group fairness comparison amongst PGFG, SLDSGNN, FairGNN, and GraphSAGE over the two different sensitive attributes. The lower value indicates less bias.}
\label{fig:group}
\vspace{-.1in}
\end{figure}

\noindent\textbf{Detailed fairness evaluation:} 
We compare the proposed PGFG model with the base GraphSAGE model, the FairGNN model, and the SLDSGNN model 
in terms of the group fairness on the two sensitive attributes, as shown in Fig.~\ref{fig:group} (a) and (b). For each sensitive group $p$ (x-axis), we report the absolute distances between RMSE achieved on each group $p$ and the overall performance across all the groups $\{p|p\in\p\}$. Here the low-income and low-education regions have larger distances because they cover a relatively small number of locations (i.e., stream segments) compared to other groups and the models in general make larger errors on these locations. 
According to Fig.~\ref{fig:group}, PGFG has substantially reduced deviations for all groups compared to all other methods, which shows the effectiveness of balancing the physical influence across different sensitive groups.

\begin{figure} [!t]
\centering
\subfigure[]{ \label{fig:c_income}{}
\includegraphics[width=0.475\columnwidth]{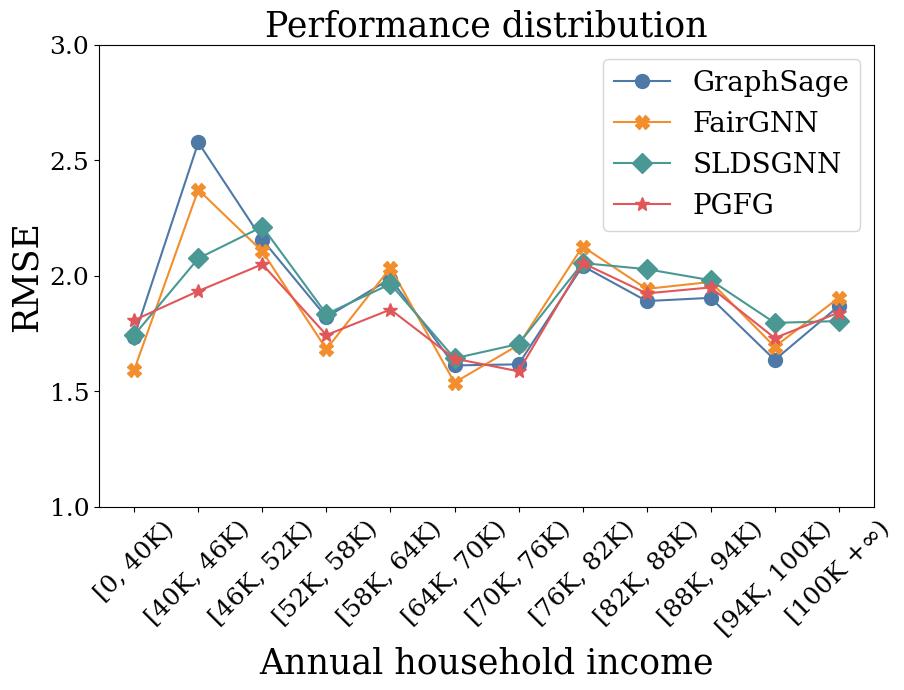}
}%\vspace{-.1in}
\subfigure[]{ \label{fig:c_edu}{}
\includegraphics[width=0.475\columnwidth]{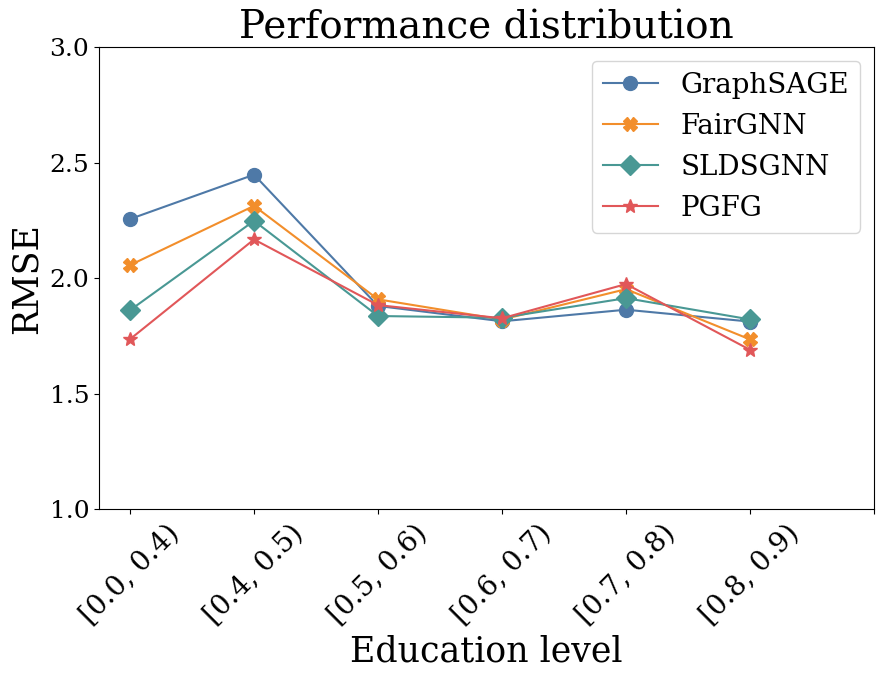}
 }
 \vspace{-.2in}
\caption{Continuous fairness comparison amongst PGFG, SLDSGNN, FairGNN, and GraphSAGE over the two different sensitive attributes. The lower value indicates less bias.}
\label{fig:continous}
\vspace{-.2in}
\end{figure}

Fig.~\ref{fig:continous} (a) and (b) present the distributions of RMSE achieved by GraphSAGE, FairGNN, and SLDSGNN within different ranges of sensitive values for the two sensitive attributes when they are considered as continuous variables. %, by the GraphSAGE model, the FairGNN model, the SLDSGNN model, and the PGFG model. 
We evaluate the model's predictive performance for stream segments within a specific range of sensitive values. It is observed that our proposed model improves mostly on poorly-performing segments in low-income and low-education communities (i.e., the left part of the distribution) while maintaining the competitive performance in well-performing segments in relatively high-income and high-education communities (the right part of the distribution).

The maps of RMSE for river segments with different income and education levels are in the techinical appendix.

% Fig.~\ref{fig:FairnessMapIncome} and Fig.~\ref{fig:FairnessMapEdu} show the maps of RMSE for a subset of segments within different income and education levels by GraphSAGE, FairGNN, and the proposed PGFG model. In particular, we present the distributions for low-, middle-, and high-income communities in the first, second, and third rows of Fig. ~\ref{fig:FairnessMapIncome}, respectively, while the first and second rows in Fig.\ref{fig:FairnessMapEdu} represent low- and high-education communities. The proposed method is shown to effectively reduce the RMSE for those segments (in red and yellow) that are poorly modeled by GraphSAGE and FairGNN. It can also be observed that PGFG still exhibits sub-optimal performance on a small number of stream segments within low-income and low-education communities. This is attributed to the influence of other local factors, such as human infrastructure, agricultural practices, and local industries, which are not accounted for by the input features used in this study, but could improve model performance. 

\section{Conclusion}

We introduce PGFG, a GNN-based method for predicting stream water temperature while ensuring fairness across regions with varying sensitive attributes. PGFG leverages physical knowledge of stream dynamics to quantify influence among graph nodes.
We then introduce an edge modifier to refine the graph structure through edge sampling based on the obtained node influence. Evaluations on a large, heterogeneous river basin show that PGFG outperforms various baselines in enforcing fairness without compromising predictive accuracy. Although PGFG is developed and evaluated in the context of stream monitoring, it is applicable to many other important societal applications, such as weather and climate modeling. Future work will explore multiple sensitive attributes from U.S. Census data and identify the attribute combinations that lead to greater spatial bias.

\section*{Acknowledgements}
This work was supported by the National Science Foundation (NSF) under grants 2239175, 2316305, 2147195, 242584, 2425845, 2430978, and 2126474, the USGS awards  G21AC10564 and G22AC00266,  the NASA grant 80NSSC24K1061,  the Google’s AI for Social Good Impact Scholars program. This research was also supported in part by the University of Pittsburgh Center for Research Computing through the resources provided. 
Any use of trade, firm, or product names is for descriptive purposes only and does not imply endorsement by the U.S. Government.

\bibliography{AAAI-2025}

\begin{thebibliography}{60}
\providecommand{\natexlab}[1]{#1}

\bibitem[{gri(2021)}]{gridMET}
 2021.
\newblock {gridMET - Climatology Lab}.
\newblock \url{http://www.climatologylab.org/gridmet.html}.

\bibitem[{Alasadi et~al.(2019)}]{alasadi2019toward}
Alasadi, J.; et~al. 2019.
\newblock Toward fairness in face matching algorithms.
\newblock In \emph{Proceedings of the 1st International Workshop on Fairness, Accountability, and Transparency in MultiMedia}, 19--25.

\bibitem[{Boyd(1996)}]{boyd1996heat}
Boyd, M.~S. 1996.
\newblock Heat source: stream, river and open channel temperature prediction.

\bibitem[{Cachay et~al.(2020)Cachay, Erickson, Bucker, Pokropek, Potosnak, Osei, and L{\"u}tjens}]{cachay2020graph}
Cachay, S.~R.; Erickson, E.; Bucker, A. F.~C.; Pokropek, E.; Potosnak, W.; Osei, S.; and L{\"u}tjens, B. 2020.
\newblock Graph Neural Networks for Improved El Ni$\backslash$\~{} no Forecasting.
\newblock \emph{arXiv preprint arXiv:2012.01598}.

\bibitem[{Carr et~al.(2013)Carr, D’Odorico, Laio, and Ridolfi}]{carr_recent_2013}
Carr, J.~A.; D’Odorico, P.; Laio, F.; and Ridolfi, L. 2013.
\newblock Recent history and geography of virtual water trade.
\newblock \emph{PloS one}, 8(2): e55825.

\bibitem[{Chen et~al.(2021)Chen, Appling, Oliver, Corson-Dosch, Read, Sadler, Zwart, and Jia}]{chen2021heterogeneous}
Chen, S.; Appling, A.; Oliver, S.; Corson-Dosch, H.; Read, J.; Sadler, J.; Zwart, J.; and Jia, X. 2021.
\newblock Heterogeneous stream-reservoir graph networks with data assimilation.
\newblock In \emph{2021 IEEE International Conference on Data Mining (ICDM)}, 1024--1029. IEEE.

\bibitem[{Chen et~al.(2022)}]{chen2022physics}
Chen, S.; et~al. 2022.
\newblock Physics-guided graph meta learning for predicting water temperature and streamflow in stream networks.
\newblock In \emph{Proceedings of the 28th ACM SIGKDD Conference on Knowledge Discovery and Data Mining}, 2752--2761.

\bibitem[{Cong et~al.(2023)Cong, Shi, Li, Yang, He, and Pei}]{cong2023fairsample}
Cong, Z.; Shi, B.; Li, S.; Yang, J.; He, Q.; and Pei, J. 2023.
\newblock FairSample: Training Fair and Accurate Graph Convolutional Neural Networks Efficiently.
\newblock \emph{IEEE Transactions on Knowledge and Data Engineering}.

\bibitem[{Dai et~al.(2021)}]{dai2021say}
Dai, E.; et~al. 2021.
\newblock Say no to the discrimination: Learning fair graph neural networks with limited sensitive attribute information.
\newblock In \emph{Proceedings of the 14th ACM International Conference on Web Search and Data Mining}, 680--688.

\bibitem[{Dong et~al.(2022)Dong, Liu, Jalaian, and Li}]{dong2022edits}
Dong, Y.; Liu, N.; Jalaian, B.; and Li, J. 2022.
\newblock Edits: Modeling and mitigating data bias for graph neural networks.
\newblock In \emph{Proceedings of the ACM Web Conference 2022}, 1259--1269.

\bibitem[{Dugdale et~al.(2017)}]{dugdale2017river}
Dugdale, S.~J.; et~al. 2017.
\newblock River temperature modelling: A review of process-based approaches and future directions.
\newblock \emph{Earth-Science Reviews}, 175: 97--113.

\bibitem[{Fan et~al.(2022)Fan, Bai, Li, Ortiz-Bobea, and Gomes}]{fan2022gnn}
Fan, J.; Bai, J.; Li, Z.; Ortiz-Bobea, A.; and Gomes, C.~P. 2022.
\newblock A GNN-RNN approach for harnessing geospatial and temporal information: application to crop yield prediction.
\newblock In \emph{Proceedings of the AAAI Conference on Artificial Intelligence}, volume~36, 11873--11881.

\bibitem[{Ge et~al.(2022)Ge, Zhao, Yu, Paul, Hu, Hsieh, and Zhang}]{ge2022toward}
Ge, Y.; Zhao, X.; Yu, L.; Paul, S.; Hu, D.; Hsieh, C.-C.; and Zhang, Y. 2022.
\newblock Toward Pareto efficient fairness-utility trade-off in recommendation through reinforcement learning.
\newblock In \emph{Proceedings of the fifteenth ACM international conference on web search and data mining}, 316--324.

\bibitem[{Hamilton et~al.(2017)}]{hamilton2017inductive}
Hamilton, W.; et~al. 2017.
\newblock Inductive representation learning on large graphs.
\newblock In \emph{Advances in neural information processing systems}, 1024--1034.

\bibitem[{Hardt et~al.(2016)}]{hardt2016equality}
Hardt, M.; et~al. 2016.
\newblock Equality of opportunity in supervised learning.
\newblock \emph{Advances in neural information processing systems}, 29.

\bibitem[{He et~al.(2022)He, Xie, Jia, Chen, Bao, Zhou, Jiang, Ghosh, and Ravirathinam}]{he2022sailing}
He, E.; Xie, Y.; Jia, X.; Chen, W.; Bao, H.; Zhou, X.; Jiang, Z.; Ghosh, R.; and Ravirathinam, P. 2022.
\newblock Sailing in the location-based fairness-bias sphere.
\newblock In \emph{Proceedings of the 30th International Conference on Advances in Geographic Information Systems}, 1--10.

\bibitem[{He et~al.(2023{\natexlab{a}})He, Xie, Liu, Chen, Jin, and Jia}]{he2023physics}
He, E.; Xie, Y.; Liu, L.; Chen, W.; Jin, Z.; and Jia, X. 2023{\natexlab{a}}.
\newblock Physics Guided Neural Networks for Time-Aware Fairness: An Application in Crop Yield Prediction.
\newblock In \emph{Proceedings of the AAAI Conference on Artificial Intelligence}, volume~37, 14223--14231.

\bibitem[{He et~al.(2024)He, Xie, Sun, Zwart, Yang, Jin, Wang, Karimi, and Jia}]{he2024fair}
He, E.; Xie, Y.; Sun, A.; Zwart, J.; Yang, J.; Jin, Z.; Wang, Y.; Karimi, H.; and Jia, X. 2024.
\newblock Fair Graph Learning Using Constraint-Aware Priority Adjustment and Graph Masking in River Networks.
\newblock In \emph{Proceedings of the AAAI Conference on Artificial Intelligence}, volume~38, 22087--22095.

\bibitem[{He et~al.(2023{\natexlab{b}})He, Zhang, Wang, Yi, Niu, and Cao}]{he2023learning}
He, H.; Zhang, Q.; Wang, S.; Yi, K.; Niu, Z.; and Cao, L. 2023{\natexlab{b}}.
\newblock Learning informative representation for fairness-aware multivariate time-series forecasting: A group-based perspective.
\newblock \emph{IEEE Transactions on Knowledge and Data Engineering}.

\bibitem[{Hipsey et~al.(2014)}]{hipsey2014glm}
Hipsey, M.; et~al. 2014.
\newblock GLM-General Lake Model: Model overview and user information.

\bibitem[{Hoekstra et~al.(2012)}]{hoekstra_water_2012}
Hoekstra, A.~Y.; et~al. 2012.
\newblock The water footprint of humanity.
\newblock \emph{Proceedings of the national academy of sciences}, 109(9): 3232--3237.

\bibitem[{Jia et~al.(2023)Jia, Chen, Zheng, Xie, Jiang, and Kalanat}]{jia2023physics}
Jia, X.; Chen, S.; Zheng, C.; Xie, Y.; Jiang, Z.; and Kalanat, N. 2023.
\newblock Physics-guided Graph Diffusion Network for Combining Heterogeneous Simulated Data: An Application in Predicting Stream Water Temperature.
\newblock In \emph{Proceedings of the 2023 SIAM International Conference on Data Mining (SDM)}, 361--369. SIAM.

\bibitem[{Jia et~al.(2019)Jia, Willard, Karpatne, Read, Zwart, Steinbach, and Kumar}]{jia2019physics}
Jia, X.; Willard, J.; Karpatne, A.; Read, J.; Zwart, J.; Steinbach, M.; and Kumar, V. 2019.
\newblock Physics Guided RNNs for Modeling Dynamical Systems: A Case Study in Simulating Lake Temperature Profiles.
\newblock In \emph{Proceedings of SIAM International Conference on Data Mining}.

\bibitem[{Jia et~al.(2021{\natexlab{a}})Jia, Xie, Li, Chen, Zwart, Sadler, Appling, Oliver, and Read}]{jia2021physics_simlr}
Jia, X.; Xie, Y.; Li, S.; Chen, S.; Zwart, J.; Sadler, J.; Appling, A.; Oliver, S.; and Read, J. 2021{\natexlab{a}}.
\newblock Physics-Guided Machine Learning from Simulation Data: An Application in Modeling Lake and River Systems.
\newblock In \emph{2021 IEEE International Conference on Data Mining (ICDM)}, 270--279. IEEE.

\bibitem[{Jia et~al.(2021{\natexlab{b}})Jia, Zwart, Sadler, Appling, Oliver, Markstrom, Willard, Xu, Steinbach, Read et~al.}]{jia2021physics_sdm}
Jia, X.; Zwart, J.; Sadler, J.; Appling, A.; Oliver, S.; Markstrom, S.; Willard, J.; Xu, S.; Steinbach, M.; Read, J.; et~al. 2021{\natexlab{b}}.
\newblock Physics-guided recurrent graph model for predicting flow and temperature in river networks.
\newblock In \emph{Proceedings of the 2021 SIAM International Conference on Data Mining (SDM)}, 612--620. SIAM.

\bibitem[{Jia et~al.(2021{\natexlab{c}})Jia, Zwart, Sadler, Appling, Oliver, Markstrom, Willard, Xu, Steinbach, Read et~al.}]{jia2021physics}
Jia, X.; Zwart, J.; Sadler, J.; Appling, A.; Oliver, S.; Markstrom, S.; Willard, J.; Xu, S.; Steinbach, M.; Read, J.; et~al. 2021{\natexlab{c}}.
\newblock Physics-guided recurrent graph model for predicting flow and temperature in river networks.
\newblock In \emph{Proceedings of the 2021 SIAM International Conference on Data Mining (SDM)}, 612--620. SIAM.

\bibitem[{Jo and Gebru(2020)}]{jo2020lessons}
Jo, E.~S.; and Gebru, T. 2020.
\newblock Lessons from archives: Strategies for collecting sociocultural data in machine learning.
\newblock In \emph{Proceedings of the 2020 Conference on Fairness, Accountability, and Transparency}, 306--316.

\bibitem[{Kamishima et~al.(2011)}]{kamishima2011fairness}
Kamishima, T.; et~al. 2011.
\newblock Fairness-aware learning through regularization approach.
\newblock In \emph{2011 IEEE 11th International Conference on Data Mining Workshops}, 643--650. IEEE.

\bibitem[{Laclau et~al.(2021)Laclau, Redko, Choudhary, and Largeron}]{laclau2021all}
Laclau, C.; Redko, I.; Choudhary, M.; and Largeron, C. 2021.
\newblock All of the fairness for edge prediction with optimal transport.
\newblock In \emph{International Conference on Artificial Intelligence and Statistics}, 1774--1782. PMLR.

\bibitem[{Lahoti et~al.(2020)Lahoti, Beutel, Chen, Lee, Prost, Thain, Wang, and Chi}]{lahoti2020fairness}
Lahoti, P.; Beutel, A.; Chen, J.; Lee, K.; Prost, F.; Thain, N.; Wang, X.; and Chi, E. 2020.
\newblock Fairness without demographics through adversarially reweighted learning.
\newblock \emph{Advances in neural information processing systems}, 33: 728--740.

\bibitem[{Lahoti et~al.(2019{\natexlab{a}})}]{lahoti2019ifair}
Lahoti, P.; et~al. 2019{\natexlab{a}}.
\newblock ifair: Learning individually fair data representations for algorithmic decision making.
\newblock In \emph{2019 ieee 35th international conference on data engineering (icde)}, 1334--1345. IEEE.

\bibitem[{Lahoti et~al.(2019{\natexlab{b}})}]{lahoti2019operationalizing}
Lahoti, P.; et~al. 2019{\natexlab{b}}.
\newblock Operationalizing individual fairness with pairwise fair representations.
\newblock \emph{arXiv preprint arXiv:1907.01439}.

\bibitem[{Li et~al.(2021)Li, Wang, Zhao, Hong, and Liu}]{li2021dyadic}
Li, P.; Wang, Y.; Zhao, H.; Hong, P.; and Liu, H. 2021.
\newblock On dyadic fairness: Exploring and mitigating bias in graph connections.
\newblock In \emph{International Conference on Learning Representations}.

\bibitem[{Liu et~al.(2024)Liu, Zhou, Guan, Peng, Xu, Tang, Zhu, Till, Jia, Jiang et~al.}]{liu2024knowledge}
Liu, L.; Zhou, W.; Guan, K.; Peng, B.; Xu, S.; Tang, J.; Zhu, Q.; Till, J.; Jia, X.; Jiang, C.; et~al. 2024.
\newblock Knowledge-guided machine learning can improve carbon cycle quantification in agroecosystems.
\newblock \emph{Nature communications}, 15(1): 357.

\bibitem[{Manson(2022)}]{manson2020ipums}
Manson, S.~M. 2022.
\newblock IPUMS national historical geographic information system: version 17.0.

\bibitem[{Markstrom(2012)}]{markstrom2012p2s}
Markstrom, S.~L. 2012.
\newblock \emph{P2S--coupled Simulation with the Precipitation-Runoff Modeling System (PRMS) and the Stream Temperature Network (SNTemp) Models}.
\newblock US Department of the Interior, US Geological Survey.

\bibitem[{Mehrabi et~al.(2021)Mehrabi, Morstatter, Saxena, Lerman, and Galstyan}]{mehrabi2021survey}
Mehrabi, N.; Morstatter, F.; Saxena, N.; Lerman, K.; and Galstyan, A. 2021.
\newblock A survey on bias and fairness in machine learning.
\newblock \emph{ACM Computing Surveys (CSUR)}, 54(6): 1--35.

\bibitem[{Moshe et~al.(2020)Moshe, Metzger, Elidan, Kratzert, Nevo, and El-Yaniv}]{moshe2020hydronets}
Moshe, Z.; Metzger, A.; Elidan, G.; Kratzert, F.; Nevo, S.; and El-Yaniv, R. 2020.
\newblock Hydronets: Leveraging river structure for hydrologic modeling.
\newblock \emph{arXiv preprint arXiv:2007.00595}.

\bibitem[{Muralidhar et~al.(2020)Muralidhar, Bu, Cao, He, Ramakrishnan, Tafti, and Karpatne}]{muralidhar2020phynet}
Muralidhar, N.; Bu, J.; Cao, Z.; He, L.; Ramakrishnan, N.; Tafti, D.; and Karpatne, A. 2020.
\newblock Phynet: Physics guided neural networks for particle drag force prediction in assembly.
\newblock In \emph{Proceedings of the 2020 SIAM International Conference on Data Mining}, 559--567. SIAM.

\bibitem[{Oliver et~al.(2021)}]{sam_release}
Oliver, S.~K.; et~al. 2021.
\newblock Predicting water temperature in the Delaware River Basin.
\newblock U.S. Geological Survey Data Release.

\bibitem[{Raissi et~al.(2017)}]{raissi2017physics1}
Raissi, M.; et~al. 2017.
\newblock Physics informed deep learning (part i): Data-driven solutions of nonlinear partial differential equations.
\newblock \emph{arXiv preprint arXiv:1711.10561}.

\bibitem[{Read et~al.(2017)}]{read2017water}
Read, E.~K.; et~al. 2017.
\newblock Water quality data for national-scale aquatic research: The Water Quality Portal.
\newblock \emph{Water Resources Research}.

\bibitem[{Regan et~al.(2018)}]{regan2018description}
Regan, R.~S.; et~al. 2018.
\newblock Description of the national hydrologic model for use with the precipitation-runoff modeling system (PRMS).
\newblock Technical report, US Geological Survey.

\bibitem[{Schwartz et~al.(2022)Schwartz, Vassilev, Greene, Perine, Burt, Hall et~al.}]{schwartz2022towards}
Schwartz, R.; Vassilev, A.; Greene, K.; Perine, L.; Burt, A.; Hall, P.; et~al. 2022.
\newblock Towards a standard for identifying and managing bias in artificial intelligence.
\newblock \emph{NIST special publication}, 1270(10.6028).

\bibitem[{Serna et~al.(2020)Serna, Morales, Fierrez, Cebrian, Obradovich, and Rahwan}]{serna2020sensitiveloss}
Serna, I.; Morales, A.; Fierrez, J.; Cebrian, M.; Obradovich, N.; and Rahwan, I. 2020.
\newblock Sensitiveloss: Improving accuracy and fairness of face representations with discrimination-aware deep learning.
\newblock \emph{arXiv preprint arXiv:2004.11246}.

\bibitem[{Steed and Caliskan(2021)}]{steed2021image}
Steed, R.; and Caliskan, A. 2021.
\newblock Image representations learned with unsupervised pre-training contain human-like biases.
\newblock In \emph{Proceedings of the 2021 ACM Conference on Fairness, Accountability, and Transparency}, 701--713.

\bibitem[{Sun et~al.(2021)Sun, Jiang, Mudunuru, and Chen}]{sun2021explore}
Sun, A.~Y.; Jiang, P.; Mudunuru, M.~K.; and Chen, X. 2021.
\newblock Explore spatio-temporal learning of large sample hydrology using graph neural networks.
\newblock \emph{Water Resources Research}, 57(12): e2021WR030394.

\bibitem[{Sweeney et~al.(2020)}]{sweeney2020reducing}
Sweeney, C.; et~al. 2020.
\newblock Reducing sentiment polarity for demographic attributes in word embeddings using adversarial learning.
\newblock In \emph{Proceedings of the 2020 Conference on Fairness, Accountability, and Transparency}, 359--368.

\bibitem[{Tang et~al.(2020)Tang, Yao, Sun, Wang, Tang, Aggarwal, Mitra, and Wang}]{tang2020investigating}
Tang, X.; Yao, H.; Sun, Y.; Wang, Y.; Tang, J.; Aggarwal, C.; Mitra, P.; and Wang, S. 2020.
\newblock Investigating and mitigating degree-related biases in graph convoltuional networks.
\newblock In \emph{Proceedings of the 29th ACM International Conference on Information \& Knowledge Management}, 1435--1444.

\bibitem[{Topp et~al.(2023)Topp, Barclay, Diaz, Sun, Jia, Lu, Sadler, and Appling}]{topp2023stream}
Topp, S.~N.; Barclay, J.; Diaz, J.; Sun, A.~Y.; Jia, X.; Lu, D.; Sadler, J.~M.; and Appling, A.~P. 2023.
\newblock Stream temperature prediction in a shifting environment: Explaining the influence of deep learning architecture.
\newblock \emph{Water Resources Research}, 59(4): e2022WR033880.

\bibitem[{USGS(2016)}]{us2016national}
USGS. 2016.
\newblock US Geological Survey. National water information system data available on the world wide web (USGS water data for the nation).
\newblock \doi{10.5066/F7P55KJN}.

\bibitem[{Wang et~al.(2020)Wang, Ma, Wang, Jin, Wang, Tang, Jia, and Yu}]{wang2020traffic}
Wang, X.; Ma, Y.; Wang, Y.; Jin, W.; Wang, X.; Tang, J.; Jia, C.; and Yu, J. 2020.
\newblock Traffic flow prediction via spatial temporal graph neural network.
\newblock In \emph{Proceedings of the web conference 2020}, 1082--1092.

\bibitem[{Willard et~al.(2022)Willard, Jia, Xu, Steinbach, and Kumar}]{willard2022integrating}
Willard, J.; Jia, X.; Xu, S.; Steinbach, M.; and Kumar, V. 2022.
\newblock Integrating scientific knowledge with machine learning for engineering and environmental systems.
\newblock \emph{ACM Computing Surveys}, 55(4): 1--37.

\bibitem[{Williamson et~al.(2015)}]{williamson2015summary}
Williamson, T.~N.; et~al. 2015.
\newblock Summary of hydrologic modeling for the Delaware River Basin using the Water Availability Tool for Environmental Resources (WATER).
\newblock Technical report, U.S. Geological Survey Scientific Investigations Report.

\bibitem[{Xie et~al.(2022)Xie, He, Jia, Chen, Skakun, Bao, Jiang, Ghosh, and Ravirathinam}]{xie2022fairness}
Xie, Y.; He, E.; Jia, X.; Chen, W.; Skakun, S.; Bao, H.; Jiang, Z.; Ghosh, R.; and Ravirathinam, P. 2022.
\newblock Fairness by “Where”: A Statistically-Robust and Model-Agnostic Bi-Level Learning Framework.
\newblock In \emph{Proceedings of the AAAI Conference on Artificial Intelligence}.

\bibitem[{Yan et~al.(2019)}]{yan2019fairst}
Yan, A.; et~al. 2019.
\newblock Fairst: Equitable spatial and temporal demand prediction for new mobility systems.
\newblock In \emph{Proceedings of the 27th ACM SIGSPATIAL International Conference on Advances in Geographic Information Systems}, 552--555.

\bibitem[{Yang et~al.(2020)Yang, Qinami, Fei-Fei, Deng, and Russakovsky}]{yang2020towards}
Yang, K.; Qinami, K.; Fei-Fei, L.; Deng, J.; and Russakovsky, O. 2020.
\newblock Towards fairer datasets: Filtering and balancing the distribution of the people subtree in the imagenet hierarchy.
\newblock In \emph{Proceedings of the 2020 Conference on Fairness, Accountability, and Transparency}, 547--558.

\bibitem[{Zafar et~al.(2017)Zafar, Valera, Gomez~Rodriguez, and Gummadi}]{zafar2017fairness}
Zafar, M.~B.; Valera, I.; Gomez~Rodriguez, M.; and Gummadi, K.~P. 2017.
\newblock Fairness beyond disparate treatment \& disparate impact: Learning classification without disparate mistreatment.
\newblock In \emph{Proceedings of the 26th international conference on world wide web}, 1171--1180.

\bibitem[{Zhang et~al.(2014)}]{zhang2014fairness}
Zhang, C.; et~al. 2014.
\newblock Fairness in multi-agent sequential decision-making.
\newblock \emph{Advances in Neural Information Processing Systems}, 27.

\bibitem[{Zhang et~al.(2021)}]{zhang2021towards}
Zhang, H.; et~al. 2021.
\newblock Towards Fair Deep Anomaly Detection.
\newblock In \emph{Proceedings of the 2021 ACM Conference on Fairness, Accountability, and Transparency}, 138--148.

\end{thebibliography}
\clearpage

\section*{Technical Appendix}
\section{Additional Details about Dataset}
The dataset we use is collected from the Delaware River Basin, which is an ecologically diverse region 
and a watershed along the east coast of the United States that provides drinking water to over 15 million people~\cite{williamson2015summary}.
The dataset used in our evaluation is from the U.S. Geological Survey's National Water Information System~\cite{us2016national} and the Water Quality Portal~\cite{read2017water}.    
Specifically, the Delaware River Basin contains 456 stream segments with input features at the daily scale from Jan 01, 1980, to Jul 31, 2020 (14,823 dates). 
The input features include meteorological variables like daily average solar radiation, air temperature, precipitation, local wind speed, date of the year, solar radiation, shade fraction, potential evapotranspiration, and the geometric features of each segment (e.g., elevation, length, slope, and width) ~\cite{sam_release}. 
Air temperature and precipitation values were derived from the GridMET gridded dataset~\cite{gridMET}. Other input features (e.g., shade fraction, potential evapotranspiration) are difficult to measure frequently, and we use values produced by the PRMS-SNTemp model~\cite{markstrom2012p2s} as its internal variables. In addition, amongst 456 segments, the number of streamflow observations only available for the 183 segments ranges from 16 to 14,774 with a total of 1,928,445 streamflow observations across all dates and segments. For those segments or specific dates where streamflow data are not available, we utilize simulated data produced by the physics-based PRMS-SNTemp model~\cite{markstrom2012p2s}. Only 30\% of the data points are replaced by simulated data, calibrated using historical measurements and statistical techniques for accuracy and reliability.

\section{Additional Details about Implementation}
This work adopts GraphSAGE as the backbone, which balances performance and runtime by sampling a small number of node neighborhoods from the adjacency matrix in large graphs.
The backbone model takes the hidden state for each segment at each time step, produced by the LSTM, as part of the input to the GNN. However, the GNN does not operate on these hidden states in isolation. For each segment, its hidden state is combined with the hidden states of its neighboring nodes, which are also processed by the LSTM for the same time step. This combined information allows the GNN to learn from both the individual segment's features and its interactions with neighbors, capturing the spatial dependencies effectively. Following the GraphSAGE principle, we randomly sample a set of nodes for training along with their neighborhoods for feature aggregation each time. The deep learning model is then trained using mini-batches of these sampled nodes, processed through LSTM and GNN layers. Finally, the model outputs temperature predictions for each node at each time step. The optimization involves backpropagation to update model parameters and minimize the loss function (e.g., MSE).

We implement the proposed method using TensorFlow 2.10 with CUDA under the environment of NVIDIA RTX A6000 GPU. Our model architecture involves a single LSTM layer with a hidden representation dimension of 20, followed by a GraphSAGE layer for aggregating these representations, and a dense layer for final output. The code for this implementation is attached in the Code \& Data Appendix to this document. We will provide pointers to our dataset once this paper is accepted and published.

In our experiments, we run all candidate methods five times with five randomly chosen seeds (1, 5, 85, 500, and 1000) for random sampling in GraphSAGE. The analysis of experimental results is conducted using the average results.

\begin{table}[]
    \scriptsize
    \centering
    \caption{The fairness and overall RMSE with a continuous sensitive attribute. The lower value indicates the better result. Bold indicates the best-performing model for a given metric and attribute.}
    \vspace{-.1in}
    \begin{tabular}{|c|cc|cccc|}
    \hline
    \multirow{3}{*}{Method} & \multicolumn{6}{c|}{Annual household income} \\ \cline{2-7} 
    & \multicolumn{2}{c|}{Discrete scenario} & \multicolumn{4}{c|}{Continuous scenario} \\ \cline{2-7} 
    & RMSE & \multicolumn{1}{c|}{Fairness} & RMSE & 1000 & 3000 & 5000 \\ \hline\hline
    STGNN & 1.775 & 0.196 & 1.775 & 4.533 & 4.411 & 3.600 \\ \hline
    PGRGrN & \textbf{1.722} & 0.171 & \textbf{1.722} & \textbf{4.432} & 3.591 & \textbf{2.501} \\ \hline
    FairFor & 1.822 & 0.136 & 1.774 & 4.935 & 4.655 & 3.557  \\ \hline
    STGNN+PGFG & 1.781 & \textbf{0.096} & 1.773 & 4.590 & \textbf{3.163} & 2.518  \\ \hline
    PGFG & 1.774 & 0.134 & 1.748 & 4.456 & 3.333 & 2.573 \\ \hline
    \end{tabular}
\label{tab:sup_performance}
\end{table} 

\begin{figure*} [!t]
\centering
\subfigure[GraphSAGE]{ \label{fig:base}{}
\includegraphics[width=0.59\columnwidth]{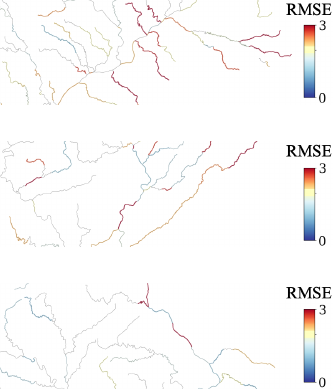}
}
\subfigure[FairGNN]{ \label{fig:dsgnn}{}
\includegraphics[width=0.59\columnwidth]{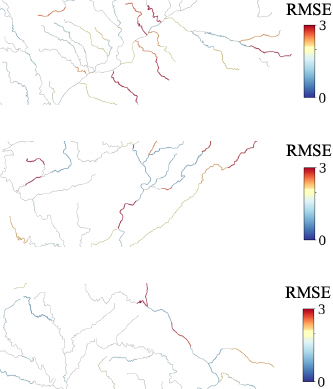}
}
\subfigure[PGFG]{ \label{fig:proposed}{}
\includegraphics[width=0.59\columnwidth]{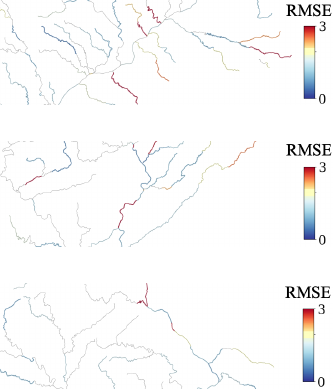}
}
\vspace{-.15in}
\caption{The distributions of predictive root mean squared error (RMSE) by (a) GraphSAGE,  (b) FairGNN, and (c) the proposed physics-guided fair graph (PGFG) model in three groups. The first row shows the performance in low-income communities, the second row shows the performance in middle-income communities, and the third row shows the performance in high-income communities. The red color indicates worse RMSE.}
\label{fig:FairnessMapIncome}
\end{figure*}

\begin{figure*} [!t]
\centering
% \hspace{-.4in}
\subfigure[GraphSAGE]{ \label{fig:base}{}
\includegraphics[width=0.48\columnwidth]{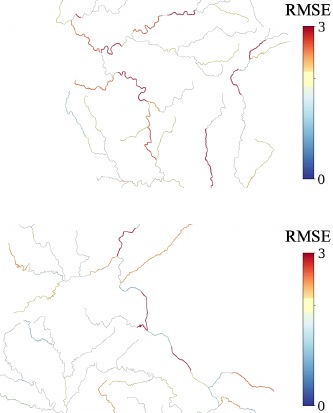}
}
\hspace{.45in}
\subfigure[FairGNN]{ \label{fig:dsgnn}{}
\includegraphics[width=0.48\columnwidth]{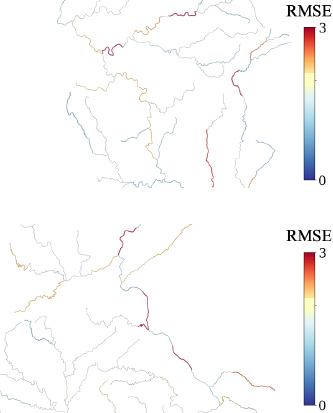}
}
\hspace{.35in}
\subfigure[PGFG]{ \label{fig:proposed}{}
\includegraphics[width=0.48\columnwidth]{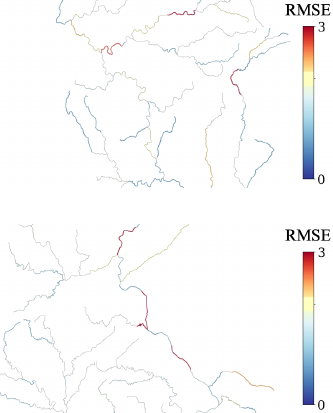}
}
\vspace{-.1in}
\caption{The distributions of predictive root mean squared error (RMSE) by (a) GraphSAGE,  (b) FairGNN, and (c) the proposed physics-guided fair graph (PGFG) model in two groups. The first row shows the performance in low-education communities, and the second row shows the performance in high-education communities. The red color indicates worse RMSE.}
\label{fig:FairnessMapEdu}
\end{figure*}

\section{More Results for Comparisons of
Candidate Methods}

In this study, we aim to predict stream water temperature and reduce model bias across locations of varying sensitive attribute values, which involves forecasting over time and space. We conduct further experiments to compare our method with two spatio-temporal graph neural networks, STGNN[1] and FairFor[2]. For STGNN (without considering fairness), we also integrate our proposed PGFG method into this STGNN framework. Additionally, physics-guided models have been applied to water stream temperature predictions. We also compare our model with a specific physics-guided model, PGRGrN~\cite{jia2021physics}, which aims to improve the local performance in segments with limited data.

Table~\ref{tab:sup_performance} presents a comprehensive overview of the performance of our proposed methods and other baselines in predicting water temperature for the Delaware River Basin. We utilize the RMSE as the evaluation metric, and the fairness performance is measured for a specific sensitive attribute, median annual household income. 

We have the following observations from Table~\ref{tab:sup_performance}:
\begin{enumerate}
    \item Compared to STGNN, other methods (e.g., PGRGrN, PGFG) achieve lower errors in fairness. However, by adopting STGNN as the backbone and integrating our method into this STGNN, the results show that we can achieve better fairness performance, demonstrating the effectiveness of our proposed approach.
    
    \item FairFor specifically focuses on group fairness. Our results indicate that PGFG achieves comparable group fairness performance to FairFor. However, FairFor tends to underperform when handling continuous sensitive attributes.

    \item PGRGrN focuses on improving the local performance in segments with limited data. Hence, our PGFG model is comparable to PGRGrN for fairness on continuous sensitive attributes. However, PGRGrN does not specifically address group fairness, which our model does, resulting in better group fairness.
\end{enumerate}

\section{More Visualizations for Comparisons of
Candidate Methods}
Fig.~\ref{fig:FairnessMapIncome} and Fig.~\ref{fig:FairnessMapEdu} show the maps of RMSE for a subset of segments within different income and education levels by GraphSAGE, FairGNN, and the proposed PGFG model. In particular, we present the distributions for low-, middle-, and high-income communities in the first, second, and third rows of Fig.~\ref{fig:FairnessMapIncome}, respectively, while the first and second rows in Fig.~\ref{fig:FairnessMapEdu} represent low- and high-education communities. The proposed method is shown to effectively reduce the RMSE for those segments (in red and yellow) that are poorly modeled by GraphSAGE and FairGNN. It can also be observed that PGFG still exhibits sub-optimal performance on a small number of stream segments within low-income and low-education communities. This is attributed to the influence of other local factors, such as human infrastructure, agricultural practices, and local industries, which are not accounted for by the input features used in this study, but could improve the model performance.

\end{document}